\documentclass[a4paper,fleqn]{cas-sc}
\usepackage{amsmath, amssymb, amsthm} 
\usepackage{graphicx} 
\usepackage{amsfonts} 
\usepackage{subcaption}
\usepackage[numbers]{natbib}
\usepackage{fancyhdr}
\usepackage{multirow}
\usepackage{caption}
\usepackage{times}
\usepackage{empheq}
\usepackage{bm}
\usepackage{algorithm}
\usepackage{algpseudocode}
\usepackage{float}
\usepackage{amsthm}  
\usepackage{xcolor}       
\newtheorem{definition}{Definition}  

\fancypagestyle{firstpagestyle}{
    \fancyhf{} 
    \fancyhead[L]{\textbf{}} 
    \fancyfoot[C]{\textbf{}} 
    \renewcommand{\headrulewidth}{0pt} 
}

\def\tsc#1{\csdef{#1}{\textsc{\lowercase{#1}}\xspace}}
\tsc{WGM}
\tsc{QE}
\tsc{EP}
\tsc{PMS}
\tsc{BEC}
\tsc{DE}
\begin{document}

\renewcommand{\headrulewidth}{0pt}

\let\WriteBookmarks\relax
\def\floatpagepagefraction{1}
\def\textpagefraction{.001}
\shorttitle{J.K. Krishnan et~al.}
\shortauthors{J.K. Krishnan et~al.}


\title [mode = title]{DAE-KAN:~A Kolmogorov-Arnold Network Model for High-Index Differential-Algebraic Equations}

\tnotetext[1]{This document reports the results of a research project partially supported by the National Natural Science Foundation of China (Grant No. 12201144), awarded to Juan Tang. The work of Mingchao Cai is also partially supported by the NIH-RCMI award (Grant No. 347U54MD013376), as well as by an affiliated project award from the Center for Equitable Artificial Intelligence and Machine Learning Systems (CEAMLS) at Morgan State University (Project ID: 02232301).}


\author[1]{Kai Luo}[type=editor,
                        auid=000,bioid=1,
                        orcid=0000-0001-0000-0000]


\affiliation[1]{organization={School of Computer Science and Cyber Engineering, Guangzhou University},
                addressline={230 Wai Huan Xi Road}, 
                city={Guangzhou},
                state={Guangdong},
                postcode={510006}, 
                country={China}}

\author[1,2]{Juan Tang}[style=chinese]
\cormark[1]

\author[3]{Mingchao Cai}[style=chinese]


\affiliation[2]{organization={Huangpu Research School of Guangzhou University},
                addressline={No. 72 Zhiming Road}, 
                city={Guangzhou},
                postcode={510555}, 
                state={Guangdong},
                country={China}}


\affiliation[3]{organization={Department of Mathematics,~Morgan State University},
                addressline={1700 East Cold Spring Ln}, 
                city={Baltimore},
                postcode={21251},             state={Maryland}, 
                country={USA}}

\author[1]{Xiaoqing Zeng}[style=chinese]

\author[1]{Manqi Xie}[style=chinese]

\author[4,5]{Ming Yan}[style=chinese]
\affiliation[4]{organization={
Institute of High Performance Computing (IHPC), Agency for Science, Technology and Research (A*STAR)},
                addressline={1 Fusionopolis Way}, 
                postcode={138632},            
                country={Singapore}}

\affiliation[5]{organization={
Centre for Frontier AI Research (CFAR), Agency for Science, Technology and Research (A*STAR)},
                addressline={1 Fusionopolis Way}, 
                postcode={138632},            
                country={Singapore}}

\cortext[cor1]{Corresponding author}


\begin{abstract}
Kolmogorov–Arnold Networks (KANs) have emerged as a promising alternative to Multi-layer Perceptrons (MLPs) due to their superior function-fitting abilities in data-driven modeling. In this paper, we propose a novel framework, DAE-KAN, for solving high-index differential-algebraic equations (DAEs) by integrating KANs with Physics-Informed Neural Networks (PINNs). This framework not only preserves the ability of traditional PINNs to model complex systems governed by physical laws but also enhances their performance by leveraging the function-fitting strengths of KANs. Numerical experiments demonstrate that for DAE systems ranging from index-1 to index-3, DAE-KAN reduces the absolute errors of both differential and algebraic variables by 1 to 2 orders of magnitude compared to traditional PINNs. To assess the effectiveness of this approach, we analyze the drift-off error and find that both PINNs and DAE-KAN outperform classical numerical methods in controlling this phenomenon. Our results highlight the potential of neural network methods, particularly DAE-KAN, in solving high-index DAEs with substantial computational accuracy and generalization, offering a promising solution for challenging partial differential-algebraic equations.

\end{abstract}


\begin{keywords}
Differential-Algebraic Equations \sep  Physics-Informed Neural Networks \sep Kolmogorov-Arnold Network \sep
\end{keywords}

\maketitle

\thispagestyle{firstpagestyle}
\pagestyle{fancy}
\fancyhead{} 
\fancyfoot{} 
\fancyfoot[C]{\thepage} 

\section{Introduction}
\label{sec1}
The development and use of Differential-Algebraic Equations (DAEs) extend over a considerable timeframe and cover diverse disciplines. In their analyses of constrained mechanical systems, Lagrange and Hamilton first introduced equations incorporating constraint conditions, which later evolved into the modern formulation of DAEs.
The system index of DAEs measures the deviation from a regular Ordinary Differential Equation (ODE) system and plays a crucial role in their analysis  \cite{Brenan1996}. High-index DAEs (index > 1) are common in various applications, such as circuit simulation  \cite{Gear1971}, chemical engineering  \cite{Biegler2007}, robotics  \cite{robotic}, and vehicle system modeling  \cite{Simeon1991}. Solving these systems is challenging due to issues like convergence, stability, and the drift-off phenomenon, where constraint errors accumulate over time  \cite{Brenan1996}. Traditional numerical methods for high-index DAEs include implicit Runge-Kutta \cite{AscherPetzold1991}, BDF methods  \cite{Cash2000}, generalized \(\alpha\)-methods  \cite{jay2009}, and Lie group methods  \cite{Liu2017, Tang2023}, but they are often limited in accuracy, especially with lower-order schemes. To address these challenges, index reduction techniques such as differentiation-based methods \cite{Gear1984}, embedded methods \cite{yang2022index}, and those designed for large-scale systems \cite{Tang2016, Tang2020} have been proposed. However, all of these techniques introduce varying levels of drift-off errors, making the selection of an appropriate method crucial for accurate solutions.

With the rapid progress in deep learning techniques, an increasing number of researchers are turning to neural networks as a tool for solving differential equations. A key approach in this area is Physics-Informed Neural Networks (PINNs), which were introduced by Raissi et al.  \cite{PINN} in 2019. The fundamental concept behind PINNs is to incorporate the physical constraints of a differential equation directly into the neural network's loss function, using residuals to merge data-driven learning with established physical principles. Since their inception, PINNs have quickly gained attention in the field of scientific AI research and have proven effective in addressing both forward and inverse problems involving Partial Differential Equations (PDEs). To enhance the performance of PINNs, Yu et al.  \cite{Yu2022} incorporated the gradient information of PDEs by adding additional gradient terms to the loss function, proposing the gradient-enhanced PINN (gPINN) framework. This approach significantly improves the accuracy and efficiency of conventional PINNs by leveraging the gradient information of PDE residuals. Additionally, to mitigate the issue of gradient backpropagation imbalance during PINN training, Wang et al.  \cite{Wang2021} analyzed the numerical stiffness-induced gradient imbalance problem and introduced an adaptive learning rate annealing algorithm. This method utilizes gradient statistics during training to balance the contributions of different terms in the composite loss function.
Recent years have seen a growing interest in developing neural network-based models for solving Differential-Algebraic Equation (DAE) systems. Kozlov et al.  \cite{Kozlov2018} introduced a semi-empirical approach combining theoretical insights with neural network training, specifically designing a model inspired by implicit Runge-Kutta methods to solve index-2 DAEs. However, their method does not extend to index-3 DAEs. Later, Moya and Lin \cite{DAE-PINN} proposed the DAE-PINN architecture, built on the PINN framework, which employs a discrete-time structure based on implicit Runge-Kutta methods. While effective for most index-1 DAEs, this model struggles with high-index systems and exhibits limited accuracy.  
To enhance neural network-based DAE solvers further, Chen et al.  \cite{Radau-PINN} developed Radau-PINN, integrating the Radau method \cite{Hairer2015} with an optimized fully connected neural network. This advancement significantly improves computational efficiency and solution accuracy, enabling the handling of index-2 DAEs. Nevertheless, even with these improvements, Radau-PINN remains ineffective for index-3 DAE systems.  

Recent advances in neural network architectures have unveiled the Kolmogorov-Arnold Network (KAN)  \cite{KAN}—a powerful alternative to conventional Multi-layer Perceptrons (MLPs). Rooted in the Kolmogorov-Arnold theorem \cite{Kolmogorov1961}, KAN replaces static linear weights with trainable B-spline-activated functions, enabling superior approximation flexibility and interpretability. Yet, despite these innovations, existing neural network methods remain limited in solving high-index DAEs, where accuracy and stability are critical challenges.
To bridge this gap, we introduce DAE-KAN, a PINN framework that synergizes the Kolmogorov-Arnold Network with the continuous PINNs methodology. Unlike prior MLP-based approaches, DAE-KAN achieves robust solutions for index-3 DAEs, a class of problems long considered intractable for neural solvers. Our experiments demonstrate that DAE-KAN not only surpasses traditional MLP-PINNs in accuracy but also addresses the limitation of numerical methods: 1. DAE-KAN has higher accuracy than traditional PINNs for solving DAEs; 2. Numerical methods are generally unable to solve high-index DAEs, and the high-index problem needs to be transformed into a problem in the form of index-1 by index reduction to solve it. However, the obtained numerical solutions can only satisfy the algebraic constraints of the index-1 formal problem, and cannot satisfy the algebraic constraints of the original DAEs. The solutions obtained by the neural network method for the index-1 formal problem can still satisfy the algebraic constraints of the high index problem; 3. Compared with traditional numerical methods, the DAE-KAN approach is capable of directly solving high-index DAEs while maintaining excellent accuracy.

The remainder of this paper is organized as follows. 
Section~\ref{sec2} begins with an analysis of the drift-off phenomenon in DAE systems, followed by a discussion of index reduction techniques.Section~\ref{sec3} We then introduce the novel DAE-KAN framework, which integrates PINNs with KANs. 
Section~\ref{sec4} presents comprehensive numerical experiments on two representative DAE systems, demonstrating the superior performance of DAE-KAN. Finally, we discuss and summarize the advantages, challenges, and potential improvements of the DAE-KAN architecture in Section~\ref{sec5}.

\section{Problem setup}
\label{sec2}
In this paper, to address two forms of DAEs—the semi-implicit form and the constrained multibody systems form—we propose the development of DAE-KAN, a deep learning framework that combines KAN \cite{KAN} with the physics-informed methodology \cite{PINN}.

A semi-implicit DAE system takes the form:
\begin{equation}
\label{2.1}
\left\{
\begin{aligned}
    u' &= f(u, z, t), \\
    0 &= g(u, z, t),
\end{aligned}
\right.\tag{2.1}
\end{equation}
where $t \in I = [0, T] \subseteq \mathbb{R}$ represents time, the prime symbol (${'}$) denotes the derivative with respect to time, \(u = u(t)\in \mathbb{R}^{n_{u}}\) denotes differential variables and \(z = z(t) \in \mathbb{R}^{n_{z}}\) denotes algebraic variables, \(f: \mathbb{R}^{n_{u}} \times \mathbb{R}^{n_{z}} \times I \to \mathbb{R}^{n_{u}} \) describes the differential equations, \(g: \mathbb{R}^{n_{u}} \times \mathbb{R}^{n_{z}} \times I \to \mathbb{R}^{n_{z}} \) describes the algebraic equations, the algebraic equation \( 0 = g(u, z, t) \) must satisfy that the Jacobian matrix of \( g \) with respect to the algebraic variables \( z\) is non-singular: \( det(\frac{\partial g}{\partial z} \neq 0)\), i.e., the matrix \(\frac{\partial g}{\partial z}\) must be invertible. 

A constrained multibody system \cite{Shabana_2013} is described by a nonlinear index-3 differential-algebraic equation system in the following form:
\begin{equation}
\label{2.2}
\left\{
\begin{aligned}
    u' &= v, \\
    M(u)v' &= f(u, v, t) - G^T(u, t)\lambda, \\
    0 &= g(u, t),
\end{aligned}
\right.
\tag{2.2}
\end{equation}
where $t \in I = [0, T] \subseteq \mathbb{R}$ represents time, $u(t) \in \mathbb{R}^{n_u}$ and $v(t) \in \mathbb{R}^{n_u}$ represent the positions and orientations of all objects and their velocities, respectively.
The vector $\bm{\lambda} \in \mathbb{R}^{n_\lambda}$ is the Lagrange multiplier vector.
The matrix $M(u) \in \mathbb{R}^{n_u \times n_u}$ is the mass matrix.
The mapping $f: \mathbb{R}^{n_u} \times \mathbb{R}^{n_u} \times I \rightarrow \mathbb{R}^{n_u}$ defines the applied forces and internal forces (excluding constraint forces),
while $g: \mathbb{R}^{n_u} \times I \rightarrow \mathbb{R}^{n_\lambda}$ (where $n_\lambda \le n_u$) defines the constraints.
The term $G(u, t)^T \lambda$ represents the constraint forces, where $G(u, t) = \partial g/\partial u \in \mathbb{R}^{n_\lambda \times n_u}$ is the Jacobian matrix of $g(u, t)$.

\subsection{Differential Index of A Differential-Algebraic Equations}
The most general form of a differential-algebraic system is that of an implicit differential-algebraic equation
\begin{equation}
\label{2.3}
    F(u'(t), u(t), t) = 0, \quad t \in I = [0,T] \subseteq \mathbb{R},\tag{2.3}
\end{equation}
where $F:\mathbb{R}^m \times \mathbb{R}^m \times I \to \mathbb{R}^m$ is sufficiently smooth, $u:I\to\mathbb{R}^m$ is the solution, and $u'$ denotes $\frac{du}{dt}$.

There are various types of DAE indices, including the differential index  \cite{Gear1984}, perturbation index \cite{Gear1984}, and solvability index  \cite{Marz2002}. The \emph{differential index} $\nu$ is the smallest integer such that the system. The definition of the differential index is as follows. 
\begin{definition}[\textbf{Index of a DAE}]
Equation \eqref{2.3} has a differentiation index = \(\nu\) if \(\nu\) is the minimal number of analytical differentiations.
\begin{equation}
\begin{aligned}
    F(u', u, t) &= 0, \\
    \frac{dF}{dt}(u', u, t) &= 0, \\
    &\vdots \\
    \frac{d^\nu F}{dt^\nu}(u', u, t) &= 0 
\end{aligned}\tag{2.4}
\end{equation}
uniquely determines $u'$ as a continuous function of $u$ and $t$, i.e., there exists a function $\mathcal{G}:\mathbb{R}^m \times I \to \mathbb{R}^m$ satisfying
\begin{equation}
    u'(t) = \mathcal{G}(u(t), t).\tag{2.5}
\end{equation}
\end{definition}

\subsection{Drift-Off Phenomenon}

The most straightforward method for index reduction is to repeatedly differentiate the algebraic constraints. However, this approach introduces errors in the constraints with each differentiation, leading to the accumulation of errors and the resulting drift-off phenomenon. To illustrate the drift-off effect, we use the classical pendulum problem as an example, demonstrating how successive differentiations of the algebraic constraints reduce the system's index. The following presents the index-3 formulation of the pendulum problem.
\begin{equation}
\label{2.6}
\left\{
\begin{aligned}
x' &= u,\\
y' &= v,\\
u' &= -\lambda x,\\
v' &= -\lambda y - 1,\\
0 &= x^2 + y^2 - 1,
\end{aligned}
\right.\tag{2.6}
\end{equation}
where \( u \) and \( v \) represent the velocities of the ball in the \( x \)- and \( y \)-directions, respectively. The final equation in \eqref{2.6} is the algebraic constraint equation. The dependent variables are \(x\), \(y\), \(u\), \(v\), \(\lambda \), with \( \lambda \) being the Lagrange multiplier. According to the algebraic constraints of equation \eqref{2.6}, we take two derivatives of \(u\) and \(v\) with respect to \(t\), respectively. In turn, we obtain the algebraic constraints \eqref{2.7} and \eqref{2.8} :
\begin{empheq}
[left=\left\{,right=\right.]{align}
\label{2.7}
0 &= xu + yv , \tag{2.7} \\
\label{2.8}
0 &= u^2 + v^2 - \lambda - y. \tag{2.8}
\end{empheq}
Replacing the algebraic equation of the system \eqref{2.6} with \eqref{2.8} transforms the problem into an index-1 problem. The algebraic equation \eqref{2.8} can be rearranged to express \( \lambda \), which helps in determining appropriate initial conditions for \( \lambda \). According to \eqref{2.8}, \(u\), \(v\), \(\lambda\) are differentiated with respect to \(t\), and after further simplification we obtain :
\begin{equation}
\left\{
\begin{aligned}
0 &= 2uu' + 2vv' - \lambda' - y',  \\
\lambda' &= -2\lambda(ux + vy) - 3v.
\end{aligned}\tag{2.9}
\right.
\end{equation}
Thus, after differentiating three times, we obtain an expression for \( \lambda' \), and the algebraic equation is transformed into an ODE. Consequently, the original system is an index-3 system.

\begin{figure}[h]  \centering\includegraphics[width=0.75\linewidth]{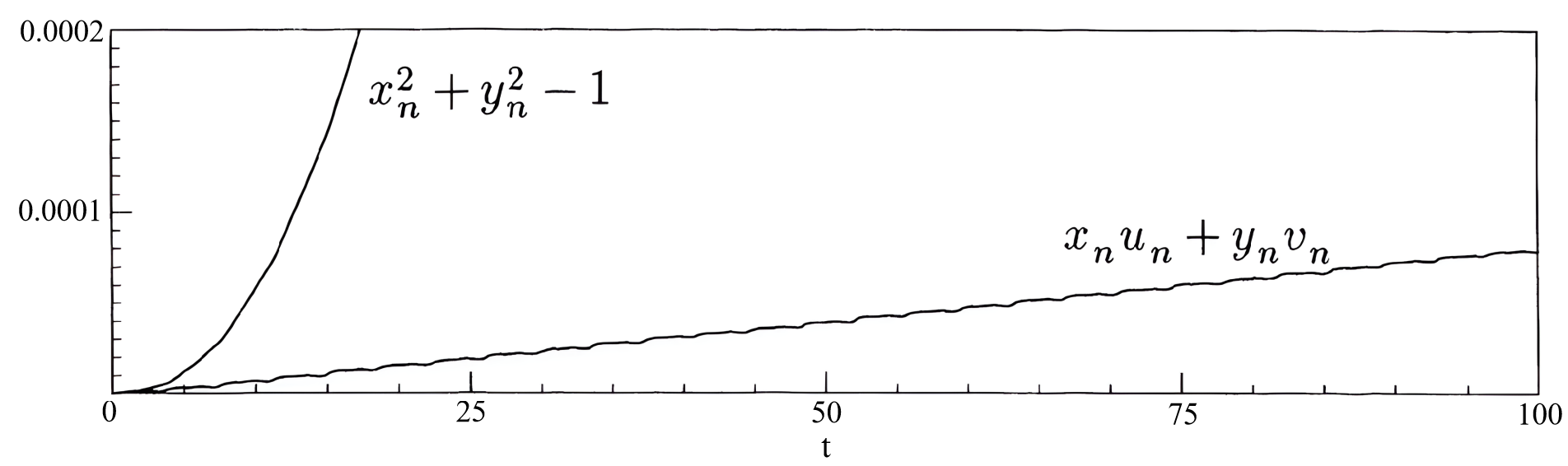}
    \caption{Error in the constraints for DOPRI5.}
    \label{fig:drift_off}
\end{figure}

We solve the index-1 form of the pendulum problem using the standard DOPRI5 solver \cite{hairer1991solving}. The initial conditions for this index-1 DAE are \( x_0 = 1, y_0 = 0, u_0 = 0, v_0 = 0 \), and Fig. \ref{fig:drift_off} demonstrates the ability of the numerical solution to preserve the constraint equations \eqref{2.7} and \( 0 = x^2 + y^2 - 1 \). As shown, the numerical solution of the index-1 system fails to accurately satisfy the algebraic constraints of the index-2 and index-3 systems, with the error progressively growing over time. It is crucial to note that our main goal is to address the simple pendulum problem in its index-3 form. In contrast, the index-1 form does not capture the constraints of the index-3 system, leading to a loss in the accuracy of the algebraic constraints. However, conventional numerical methods are often unable to solve high-index DAEs directly and must first reduce them to lower-index forms. This transformation process tends to increase the error in the algebraic constraints, which remains a significant challenge for traditional methods when dealing with high-index DAEs.

\section{Proposed framework: DAE-KAN}
\label{sec3}
In this section, we provide a concise overview of the PINN method and introduce the fundamental concepts of Kolmogorov-Arnold networks. Additionally, we conclude this section by presenting our DAE-KAN framework.
\subsection{The brief review of PINNs}
Physics-Informed Neural Networks (PINNs) \cite{PINN} represent an innovative approach initially proposed for solving partial differential equations (PDEs), integrating physical laws as constraints into data-driven neural network frameworks. The core principle of PINNs lies in embedding the governing equations of the physical system into the neural network's loss function. This formulation enables the neural network to efficiently learn the underlying physical laws from sparse and noisy data, enhancing both accuracy and generalization in solving equations.
\begin{figure}[h]  
    \centering  
    \includegraphics[width=0.8\linewidth]{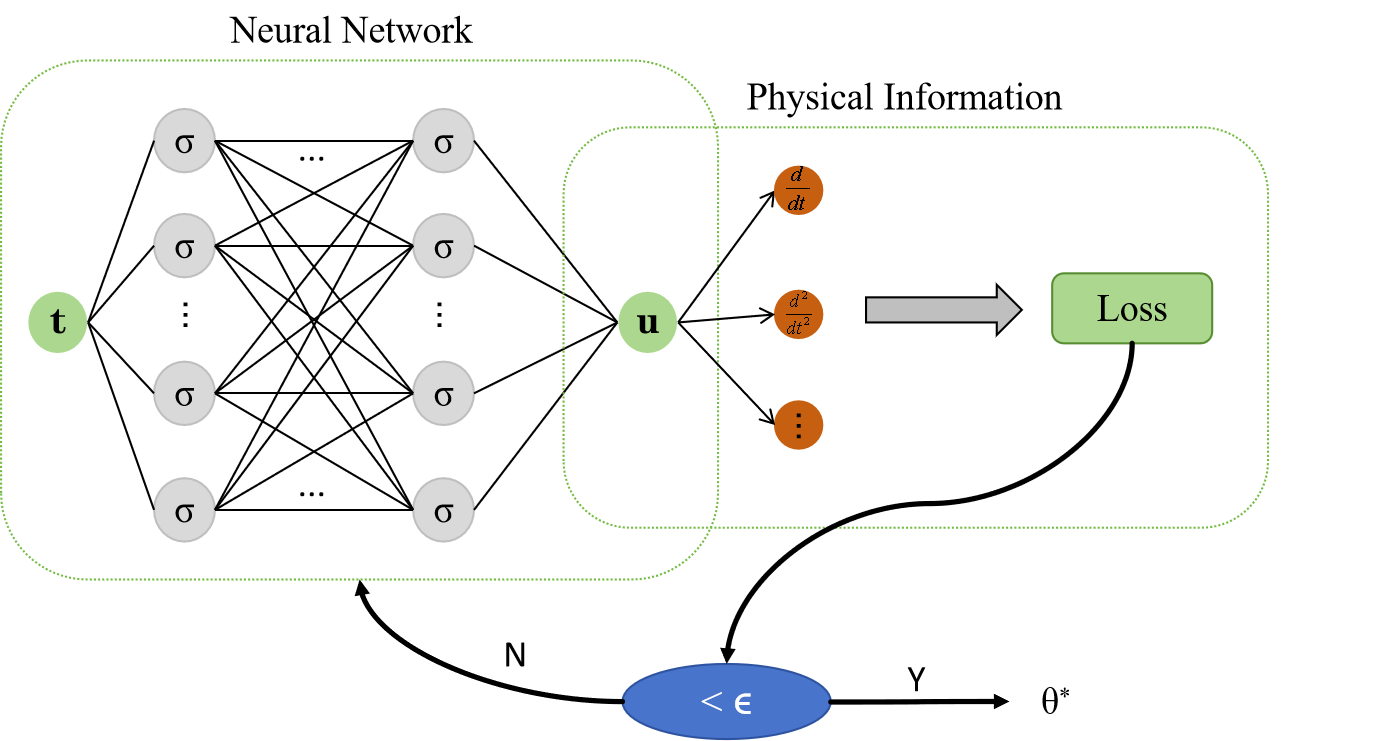}  
    \caption{Conventional physical information neural networks with MLP.}  
    \label{fig:MLP}  
\end{figure}
In the past, the MLPs employed in PINNs were primarily grounded in the Universal Approximation Theorem (UAT) \cite{Hornik1989}, and incorporated the physical information contained in equations into the neural network model to construct a neural network approximation of the solution:  
\begin{align}
    {u^{\text{PINN}}(t,\bm{\theta})} \approx u(t),\tag{3.1}
\end{align}
where \( \bf{\theta} \) is the neural network parameter and \(u(t)\) denote the exact solution of the equation at time \(t\), \(u^{\text{PINN}}(t,\bm{\theta})\) is the predicted solution of the neural network at time \(t\), which aims to get \(u^{\text{PINN}}(t,\bm{\theta})\)  close enough to \(u(t)\).  Fig. \ref{fig:MLP} illustrates the framework diagram of the MLP-based PINNs.
In traditional PINNs, the neural network \(u^{\text{PINN}}(t,\bm{\theta})\) is typically constructed using a MLP, consisting of an input layer, several fully connected hidden layers (with each neuron having a fixed nonlinear activation function), and an output layer. The transformation in each layer involves a weight matrix \( \bm{W}_l \) and an activation function \( \bm{\sigma}_l \). In Fig. \ref{fig:MLP} ,  \(\bf{t}\) represents the input vector of the neural network and the following gives the output:
\begin{align}
     \text{MLP}(\mathbf{t}) = (\bm{W}_{L-1} \circ \bm{\sigma}_{L-1} \circ \bm{W}_{L-2} \circ \bm{\sigma}_{L-2} \circ \cdots \circ \bm{W}_1 \circ \bm{\sigma}_1 \circ \bm{W}_0)(\mathbf{t}). \tag{3.2}
\end{align} 
The training process is trying to minimize the loss, where the loss usually consists of three parts: the initial condition loss, the boundary condition loss, and the residual loss of the governing equation. Through iterative optimization processes, the optimal neural network model can be effectively achieved.

\subsection{The DAE-KAN framework}
In contrast to MLP based on the Universal Approximation Theorem \cite{Hornik1989}, KANs are based on the Kolmogorov-Arnold Representation Theorem (KAT) \cite{Kolmogorov1961}, which provides a theoretical method for representing continuous functions. The theorem states that if \( f \) is a continuous function defined on a bounded domain, then \( f \) can be written as a finite combination of one-dimensional continuous functions with binary operations of addition, such that:
\begin{align}
\label{3.3}
f(\bm{x}) = f(x_1,x_2,\cdots,x_k)=\sum_{q=1}^{2k+1}\Phi_q(\sum_{p=1}^{k}\phi_{q,p}(x_p)),\tag{3.3}
\end{align}
where \( x_p \) is the \( p^{th} \) component of vector \(\bm{x}\), with \( p \) ranging from 1 to \( n \) (the dimension of the input vector); \( q \) is the index (used to iterate over each component of the outer function \( \Phi \)); \( \phi_{q,p} \) is the inner function, \( \phi_{q,p}: [0,1] \to \mathbb{R} \), which handles the \( p^{th} \) component of the input vector \(\bm{x}\) and contributes one term to the summation of the \( q^{th} \) outer function; \( \Phi_q \) is the outer function, \( \Phi_q \to \mathbb{R} \), used to aggregate the outputs of the inner functions \( \phi_{q,p} \). 
The theorem \cite{Kolmogorov1961} shows that learning a high-dimensional function \( f \) can be reduced to learning the combination of one-dimensional functions \( \phi_{q,p} \) and \( \Phi_q \). This provides a direct theoretical basis for the KAN network structure. 

\begin{figure}[h]  
    \centering  
    \includegraphics[width=0.5\linewidth]{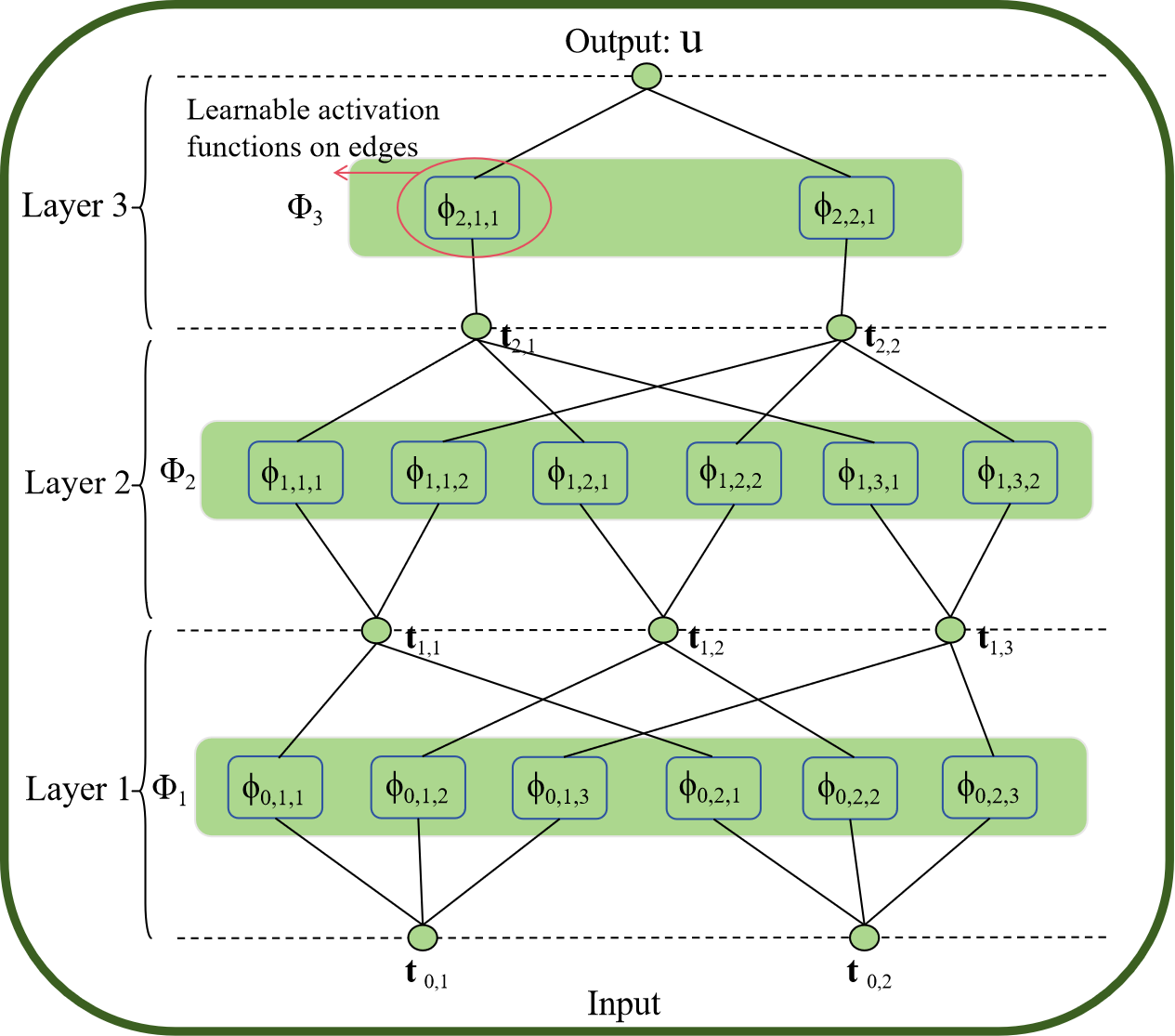}  
    \caption{Illustration of a 3-layer KAN network of the shape [2,3,2,1].}  
    \label{three_layer_KAN}  
\end{figure}

The KAN solution aims to approximate the exact solution using the KAN structure. Specifically, we have
\[
u^{\text{KAN}}(t,\bm{\theta}) \approx u(t), \tag{3.4}
\]
where \( \bm{\theta} \) denotes the set of trainable weights in the KAN network, and \( u(t) \) represents the exact solution of the DAE system. Our goal is to ensure that \( u^{\text{KAN}}(t,\bm{\theta}) \) closely approximates \( u(t) \).
Leveraging the hierarchical structure established in Theorem \eqref{3.3}, Liu et al. \cite{KAN} extended the KAT to a more general KAN architecture with arbitrary width and \(L\) layers.
\begin{align}    
\mathbf{u} &= \text{KAN}(\mathbf{t}) = (\Phi_{L-1}\circ\Phi_{L-2}\circ\cdots\circ\Phi_{1}\circ\Phi_0)(\mathbf{t}),\tag{3.5}\\
 \Phi_l &= 
\begin{pmatrix}
\phi_{l,1,1}(\cdot)\qquad\phi_{l,1,2}(\cdot) \qquad\cdots \qquad\phi_{l,1,n_l}(\cdot)\\
\phi_{l,2,1}(\cdot)\qquad\phi_{l,2,2}(\cdot) \qquad\cdots \qquad\phi_{l,2,n_l}(\cdot)\\
\vdots\qquad\qquad\qquad\vdots\qquad\qquad\qquad\qquad\vdots\\
\phi_{l,n_{l+1},1}(\cdot)\qquad\phi_{l,n_{l+1},2}(\cdot) \quad\cdots \qquad\phi_{l,n_{l+1},n_l}(\cdot)
\end{pmatrix},\tag{3.6}
\end{align}
As in Fig. \ref{three_layer_KAN}, \(\bf{t}\) is an input vector, \( \phi_{l,i,j} \) has trainable parameters, which is the
activation function that connects the \(i^{th}\) neuron in the \(l^{th}\) layer and the \(j^{th}\) neuron in the \((l + 1)^{th}\) layer, \( l \) is the layer index, and \( n_l \) and \( n_{l+1} \) represent the number of nodes in layers \( l \) and \( l+1 \), respectively. Here, we use the B-spline as the grid basis function proposed by Liu et al. \cite{KAN}.
\begin{align}
\phi_{l,i,j}(\mathbf{t}_{l, i}) = w_{l,i,j}\cdot(b(\mathbf{t}_{l, i})+spline(\mathbf{t}_{l, i})),\tag{3.7}
\end{align}
where \( w_{l,i,j} \) is the weight factor controlling the overall amplitude of the activation function, and \( b(t) \) is the basis function, which can be set as:
\begin{align}
b(\mathbf{t}_{l, i}) = silu(\mathbf{t}_{l, i}) = \frac{\mathbf{t}_{l, i}}{1+e^{-\mathbf{t}_{l, i}}},\tag{3.8}
\end{align}
the \( spline(\mathbf{t}_{l, i}) \) is a spline function, which can be parameterized as a linear combination of B-splines:
\begin{align}
spline(\mathbf{t}_{l, i}) = \sum_s c_s  B_s(\mathbf{t}_{l, i}),\tag{3.9}
\end{align}
where \(\mathbf{t}_{l, i}\) denote the \(i^{th}\) input of the \(l^{th}\) layer KAN and \( B_s(\mathbf{t}_{l, i}) \) is the B-spline function. During training, \( spline(\cdot) \) and \( w_{l,i,j} \) are trainable, and we initialize the B-spline coefficients \( c_s \sim N(0, 0.1^2) \) and the weights \( w_{l,i,j} \) using Xavier initialization. The shape of a KAN is represented by an integer array: \([n_0,n_1,\cdots,n_L]\),  where \(n_i\) is the number of nodes in the \(i^{th}\) layer of the computational graph. We denote the \(i^{th}\) neuron in the \(l^{th}\) layer by \((l, i)\), and the activation value of the \((l, i)\)-neuron by \(\mathbf{t}_{l, i}\). For example, Fig. \ref{three_layer_KAN} shows a three-layer KAN network. Unlike MLP, the B-spline functions in Fig. \ref{three_layer_KAN} are placed on the edges rather than at the nodes. It is worth noting that various grid basis functions can be used here. For example, Li et al. \cite{li2024kolmogorov} use Gaussian radial basis functions as the grid basis functions.
In the DAE problem \eqref{2.1} and \eqref{2.2}, we define the loss function as:
\begin{align}
\label{2.16}
Loss(\theta) = MSE_F + MSE_i,\tag{3.10}
\end{align}
where:
\begin{align}
MSE_F &= \frac{1}{N_F}\sum_{n=1}^{N_F} |{F(u'^{\text{KAN}}(t_F^{n}), u^{\text{KAN}}(t_F^{n}), t_F^{n})}|^2, \tag{3.11}  \\ 
MSE_i &= \frac{1}{N_i}\sum_{n=1}^{N_i} |{u^{\text{KAN}}(t_i^{n})}-{u^n)}|^2.\tag{3.12}
\end{align}
Here, loss \(MSE_i\) corresponds to the initial data, while \(MSE_F\) enforces the structure imposed by equation \eqref{2.1} or \eqref{2.2} at a finite set of collocation data points. The loss \(MSE_i\) is calculated over \( N_i \) initial data points, and \( MSE_F \) is calculated over \( N_F \) collocation points. \(u'^{\text{KAN}}(t_F^{n})\), \(u^{\text{KAN}}(t_F^{n})\) and \(u^{\text{KAN}}(t_i^{n})\) are outputs of the KAN networks, while \(u^n\) is the exact value of the \(n^{th}\) data point.
\begin{figure}[h]
    \centering
    \includegraphics[width=1\linewidth]{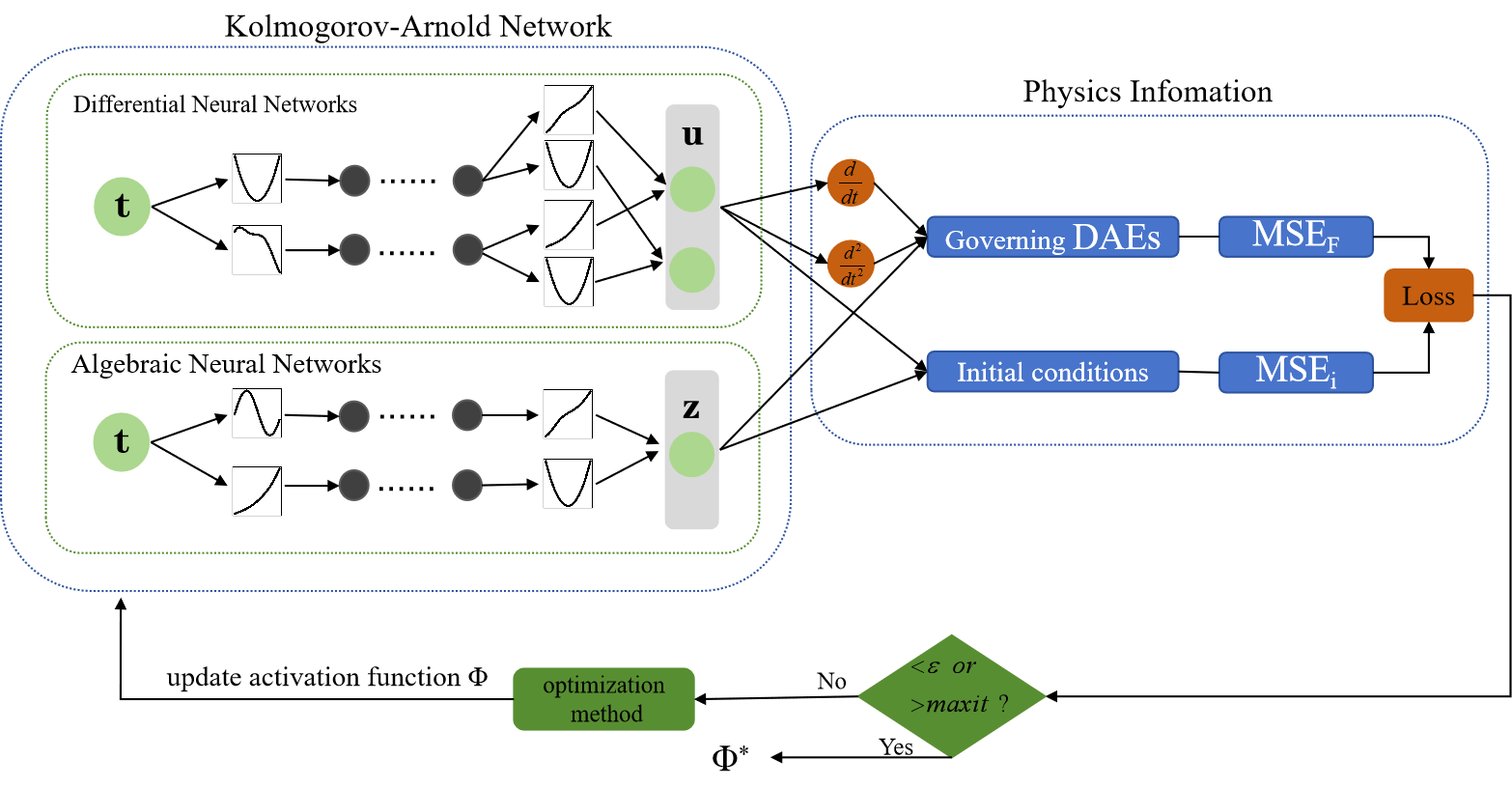}
    \caption{Diagram of the DAE-KAN network structure.}
    \label{fig:DAE-KAN}
\end{figure}
Fig. \ref{fig:DAE-KAN} describes the overall framework of the DAE-KAN network, where the differential neural network predicts the differential variables, and the algebraic neural network predicts the algebraic variables. Using two KAN networks to separately predict the differential and algebraic variables helps the network better learn the physical information of the DAEs and obtain better predictive accuracy. The input to the KAN network is a training dataset consisting of initial and residual points, where residual points are generally uniformly sampled within the time domain. Each layer of the network consists of trainable B-spline activation functions, parameterized by the learnable coefficients of the local B-spline basis functions. By stacking multiple KAN layers, the KAN network can flexibly approximate complex high-dimensional functions. The outputs of the differential and algebraic neural networks (differential variables \( \bf{u} \) and algebraic variables \( \bf{z} \)) are automatically differentiated to satisfy the physical constraints of the DAE systems. By iteratively training and optimizing the activation functions \( \Phi \) to minimize the loss function in equation \eqref{2.16}, the DAE systems physical information is embedded in the Kolmogorov-Arnold neural network architecture.Thus, by minimizing the total loss function over the KAN parameters, we obtain the optimal KAN:
\begin{align}
\Phi^* = \underset{\Phi}{argmin}(MSE_F + MSE_i),\tag{3.13}
\end{align}
To solve the optimization problem above, we can use LBFGS or Adam optimizers, and ultimately, we will obtain a neural network model containing the physical information of the DAE systems.The proposed DAE-KAN for DAEs can be summarized in \(\mathbf{Algorithm}\) \(\mathbf{1}\).
\begin{algorithm}
\caption{DAE-KAN for capturing differential-algebraic equations}
\label{alg:DAE-KAN}
\begin{algorithmic}[1]
\Require DAE systems training dataset generated by time domain simulation; DAE systems parameters (e.g., $\bf{u}$, $\bf{z}$ in \eqref{2.1})
\Ensure KAN parameters  
\State Initialize KAN parameters: $\{\Phi_l\}_{l=1}^{L}$
\State Specify the loss function as equation (\ref{2.16})
\State Specify the initial training data points: $\{(t_i^n, \bm{u}^n)\}_{n=1}^{N_i}$, and collocation training points: $\{(t_F^n)\}_{n=1}^{N_F}$
\State Set the maximum number of training steps $N$, and learning rate
\While{$n_{\text{iter}} < N$}
    \State Forward pass of KAN to calculate all $\bm{u}(t_i^n)$
    \State Calculate $MSE_i$ based on the output of KAN and the initial value
    \State Calculate $MSE_F$ based on the output of KAN and the DAEs given in equation \eqref{2.1} or \eqref{2.2}
    \State Find the best KAN parameters to minimize the loss function using the LBFGS optimizer
    \If{$n_{\text{iter}} \mod 10 == 0$}
        \State Evaluate the performance of the DAE-KAN loss function
    \EndIf
    \State $n_{\text{iter}} \gets n_{\text{iter}} + 1$ \Comment{Increment iteration counter}
\EndWhile
\end{algorithmic}
\end{algorithm}
\section{Numerical Experiments}
\label{sec4}
In this section, we validate the effectiveness of the DAE-KAN model in solving differential-algebraic equations using two different DAE systems. We use the absolute error (\(\text{\text{AE}}\)),
\(
\text{AE} = \sum_{i=1}^{N} |u_{true}^{i} - u_{pre}^{i}|,
\)
to measure the accuracy of the model’s predictions. Additionally, we introduce the \(l^2\) relative error(\(\text{RE}\)),
\begin{align}
\text{RE} = \frac{\sqrt{\sum_{i=1}^{N} |u_{true}^{i} - u_{pre}^{i}|^2}}{\sqrt{\sum_{i=1}^{N} |u_{true}^{i}|^2}},\tag{4.1}
\end{align}
to quantify the overall discrepancy between the predicted solution and the exact solution. Here, $u_{true}^{i}$ represents the exact solution, $u_{pre}^{i}$ represents the neural network's predicted solution, and $N$ denotes the number of test points.

\subsection{Constrained Motion of a Particle on a Circular Orbit}
Consider the following nonlinear index-3 form of a DAE, which describes the constrained motion of a particle on a circular trajectory \cite{Benhammouda2015}:
\begin{equation}
\label{4.2}
\left\{
\begin{aligned}  
    u_1' &= z_1,\\
    u_2' &= z_2,\\
    z_1' &= 2u_2 - 2u_2^3 - u_1 \lambda, \\
    z_2' &= 2u_1 - 2u_1^3 - u_2 \lambda, \\
            0 &= u_1^2 + u_2^2 - 1, \quad t \geq 0.
\end{aligned}
\right.\tag{4.2}
\end{equation}
\begin{figure}[h]
    \centering
    \includegraphics[width=1\linewidth]{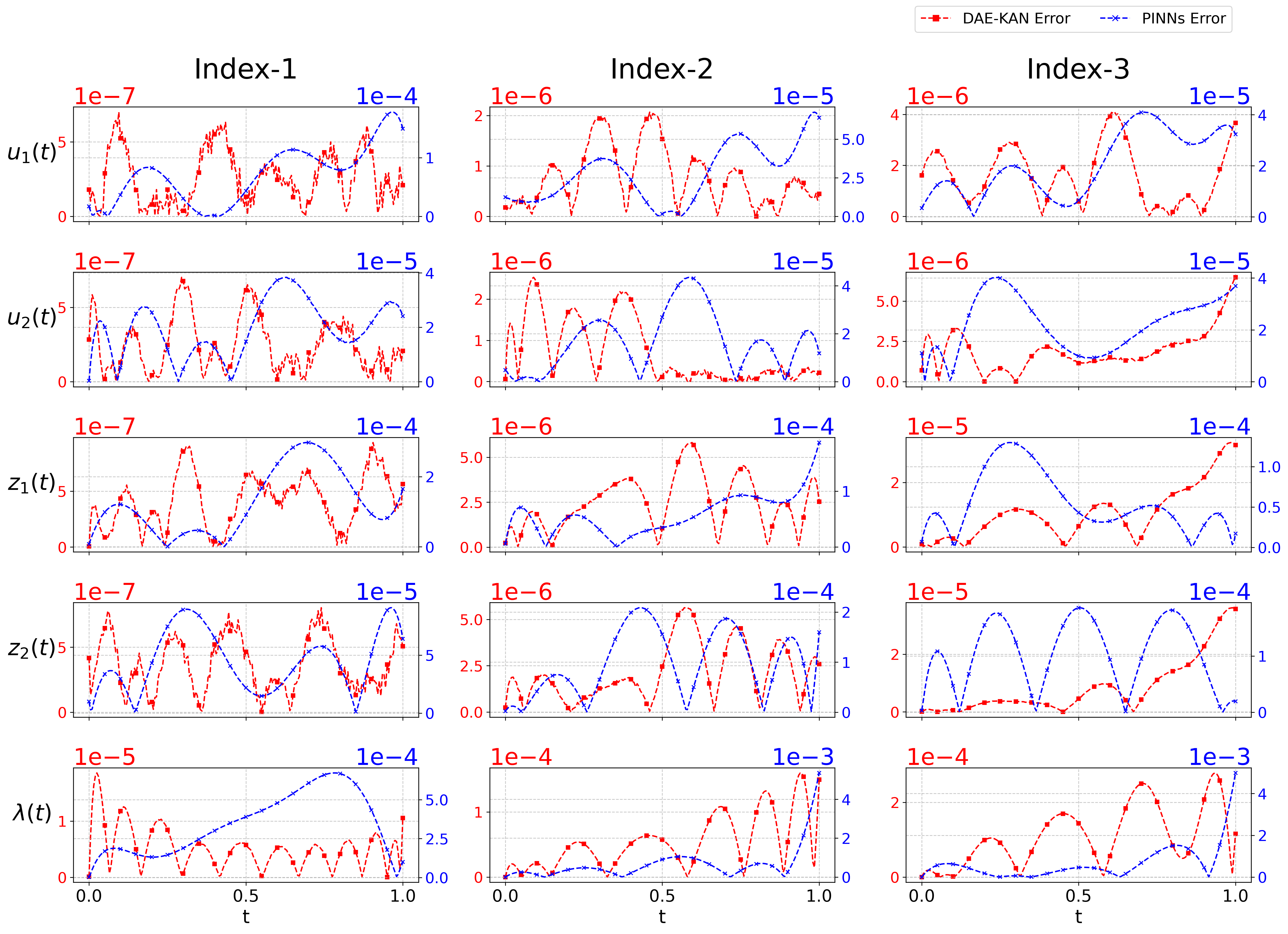}
    \caption{The \(\text{AE}\) curves of the predicted solution and the exact solution obtained by using the traditional PINNs and DAE-KAN to solve the corresponding index-1, index-2, and index-3 forms of equation \eqref{4.2}, respectively. The green line represents DAE-KAN, and the blue line represents PINNs.}
    \label{fig:particle_Absolute_errors}
\end{figure}
Differentiating the algebraic constraint equation once and simplifying, we obtain:
\begin{align}
\label{4.3}
u_1z_1 + u_2z_2 = 0, \tag{4.3}
\end{align}
thus, the first four equations of \eqref{4.2} along with \eqref{4.3} constitute the index-2 form of this example. Similarly, differentiating \eqref{4.3} once more and simplifying, we get:
\begin{align}
\label{4.4}
z_1^2 + z_2^2 + 2u_1u_2(2 - u_2^2 - u_1^2) - \lambda (u_2^2 + u_1^2) = 0,\tag{4.4}
\end{align}
replacing the algebraic constraint equation in \eqref{4.2} with  \eqref{4.4}, we obtain the index-1 form of this system.
The DAE systems \eqref{4.2} is provided with the following consistent initial conditions:
\begin{align}
u_1(0) = 1, \quad z_1(0) = 0, \quad u_2(0) = 0, \quad z_2(0) = 1, \quad \lambda(0) = 1.\tag{4.5}
\end{align}

The exact solution of this DAE systems is given by:
\begin{align}
u_1(t) = \cos(t), \quad u_2(t) = \sin(t), \quad z_1(t) = -\sin(t), \quad z_2(t) = \cos(t), \quad \lambda(t) = 1 + \sin(2t).\tag{4.6}
\end{align}
The differential KAN network in the DAE-KAN model is designed as a three-layer KAN network with the structure \([1,5,5,4]\), while the algebraic KAN network is also a three-layer KAN with the structure \([1,5,5,1]\). Following the loss function outlined in \eqref{2.16}, we set the number of initial points to 1 (\(N_i = 1\)) and the number of residual points to 200 (\(N_F = 200\)), with these 200 points uniformly distributed over the interval \(t \in [0,1]\). This results in a training dataset containing 201 points, including both initial and residual points. The training process is carried out using the LBFGS algorithm for 24,000 epochs. The absolute error between the predicted solution and the exact solution is shown by the green curve in Fig. \ref{fig:particle_Absolute_errors}.

This study further compares the traditional PINNs model \cite{PINN} with the DAE-KAN model in solving and predicting Equation \eqref{4.2} for three different formulations: index-1, index-2, and index-3. The traditional PINNs model consists of a deep neural network with five hidden layers, each containing 60 neurons, and uses the \(Tanh\) activation function. The Absolute Error (AE) curves for both models are shown in Fig. \ref{fig:particle_Absolute_errors}, where the blue curve represents the AE for the traditional PINNs model, and the red curve represents that for the DAE-KAN model. As observed in Fig. \ref{fig:particle_Absolute_errors}, when directly solving the differential variables \(u_1\), \(u_2\), \(z_1\), \(z_2\) in the index-3 formulation of Equation \eqref{4.2}, the traditional PINNs model achieves an AE precision of \(10^{-4} \sim 10^{-5}\), while the algebraic variable \(\lambda\) has an AE precision of \(10^{-3}\). In contrast, the DAE-KAN model achieves an AE precision of \(10^{-5} \sim 10^{-6}\) for the differential variables, improving accuracy by one order of magnitude. Similarly, the AE precision for the algebraic variable \(\lambda\) is improved by an order of magnitude, reaching \(10^{-4}\).

\begin{table}[width=.9\linewidth,cols=6,pos=h]
\renewcommand{\arraystretch}{1.5}
\caption{Comparison of \(\text{RE}\) for different models and variables.}\label{tb1}
\begin{tabular*}{\tblwidth}{@{} LLLLLL@{} }
\toprule
Model & $u_1(t)$ & $u_2(t)$ & $z_1(t)$ & $z_2(t)$ & $\lambda(t)$ \\
\midrule
DAE-KAN(index-3) & \textbf{2.27e-06} & \textbf{4.29e-06} & \textbf{2.52e-05} & \textbf{1.46e-05} & \textbf{7.69e-05} \\
PINNs(index-3) & 2.81e-05 & 4.67e-05 & 1.25e-04 & 1.40e-04 & 5.80e-04 \\
DAE-KAN(index-2) & \textbf{1.13e-06} & \textbf{1.96e-06} & \textbf{5.47e-06} & \textbf{3.04e-06} & \textbf{3.80e-05} \\
PINNs(index-2) & 3.83e-05 & 3.94e-05 & 1.29e-04 & 1.36e-04 & 6.06e-04 \\
DAE-KAN(index-1) & \textbf{3.69e-07} & \textbf{5.83e-07} & \textbf{9.09e-07} & \textbf{4.95e-07} & \textbf{3.64e-06} \\
PINNs(index-1) & 4.29e-05 & 7.93e-05 & 1.92e-04 & 6.90e-05 & 9.38e-05 \\
\bottomrule
\end{tabular*}
\end{table}
By replacing the algebraic equation in \eqref{4.2} with \eqref{4.3}, we obtain the index-2 formulation. The traditional PINNs model achieves an \(\text{AE}\) precision of \(10^{-4} \sim 10^{-5}\) for the differential variables \(u_1\), \(u_2\), \(z_1\), \(z_2\) and \(10^{-3}\) for the algebraic variable \(\lambda\). In contrast, the DAE-KAN model achieves an \(\text{AE}\) precision of \(10^{-6}\) for the differential variables, where \(u_1, u_2\) are improved by an order of magnitude, and \(z_1\), \(z_2\) are improved by two orders of magnitude. The algebraic variable \(\lambda\) also improves by an order of magnitude, reaching \(10^{-4}\). Similarly, by replacing the algebraic equation in Equation \eqref{4.2} with \eqref{4.4}, we obtain the index-1 formulation. In this case, the traditional PINNs model achieves an \(\text{AE}\) precision of \(10^{-4} \sim 10^{-5}\) for the differential variables \(u_1\), \(u_2\), \(z_1\), \(z_2\) and \(10^{-4}\) for the algebraic variable \(\lambda\). In contrast, the DAE-KAN model achieves an \(\text{AE}\) precision of \(10^{-7}\) for the differential variables, where \(u_2\), \(z_2\) improve by two orders of magnitude, and \(u_1\), \(z_1\) improve by three orders of magnitude. The algebraic variable \(\lambda\) also improves by an order of magnitude, reaching \(10^{-5}\).

According to Table \ref{tb1}, we compare the \(\text{RE}\) between the predicted and exact solutions for the three formulations of \eqref{4.2} using both models. In the index-3 formulation, the maximum overall \(\text{RE}\) for the differential variable predictions using the DAE-KAN model is \(2.52\times10^{-5}\), outperforming the traditional PINNs model by an order of magnitude. The DAE-KAN model also provides superior predictions for the algebraic variables. In the index-2 and index-1 formulations, the overall maximum \(\text{RE}\) of the DAE-KAN model remains better than that of the traditional PINNs model, with some variables exhibiting improvements of up to two orders of magnitude.

Overall, the DAE-KAN model demonstrates lower \(\text{AE}\) and \(\text{RE}\)  compared to the traditional PINNs model in solving the equations in all three formulations (index-1, index-2, and index-3). This is primarily due to the use of the KAN instead of the MLP used in traditional PINNs, highlighting the superiority of KAN in high-dimensional function approximation.

\begin{figure}[h]
    \centering
    \includegraphics[width=1\linewidth]{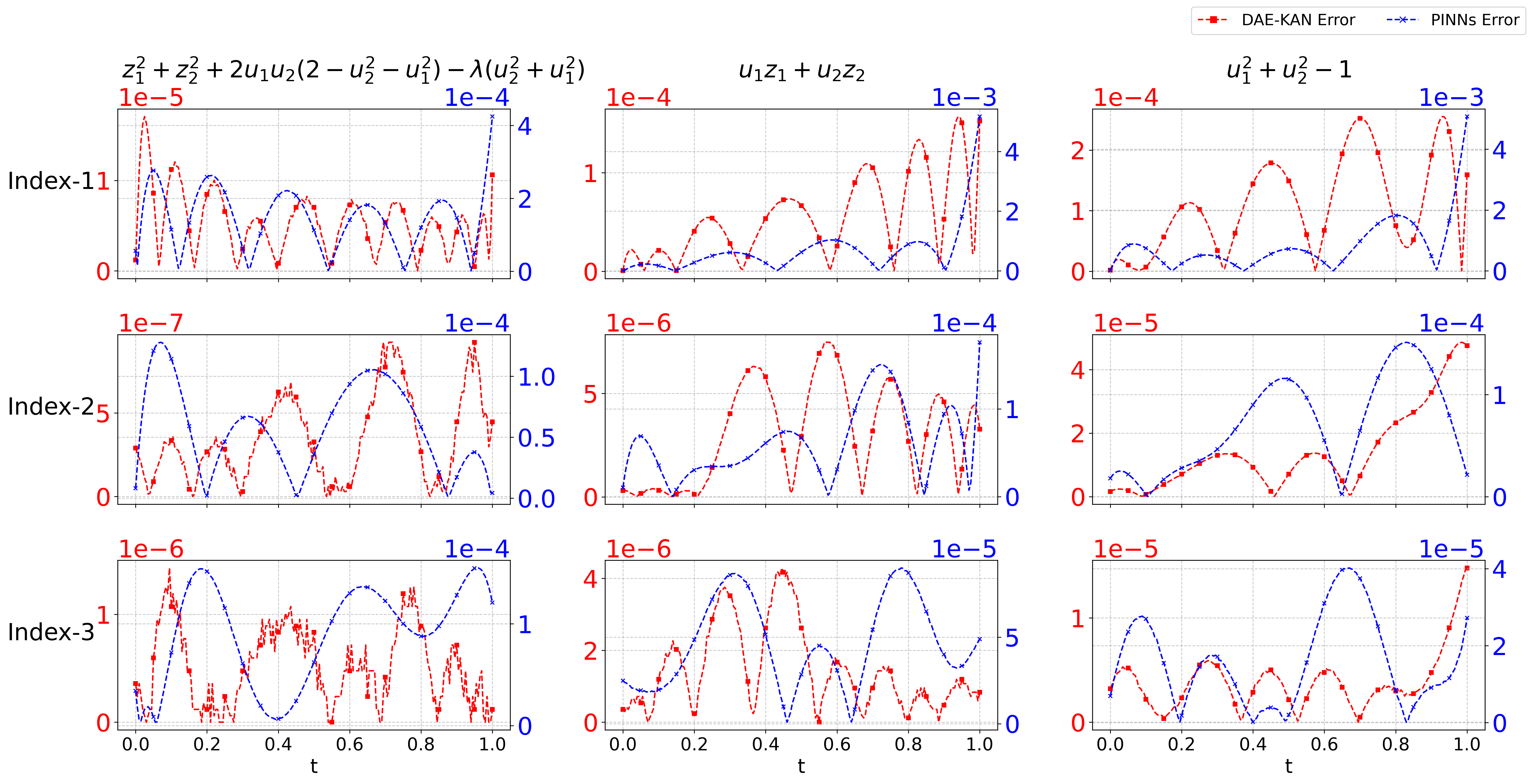}
    \caption{The predicted solutions in the form of \eqref{4.2} index-1, index-2, and index-3 obtained by PINNs and DAE-KAN are respectively inserted into the constraint equations in the form of \eqref{4.2} index-1, index-2, and index-3. That is, the \eqref{4.4}, \eqref{4.3}, and the constraint equations of \eqref{4.2}, resulting in the drift-off error curve. The green line represents the drift-off error of DAE-KAN, and the blue line represents the drift-off error of PINNs.}
    \label{fig:particle_drift_off}
\end{figure}
Most high-index DAEs cannot be directly solved using traditional numerical methods, as they require transformation into index-1 problems through index reduction. However, numerical solutions obtained by solving the index-1 form often fail to satisfy the algebraic constraints of high-index problems, leading to drift-off errors, as illustrated in Fig. \ref{fig:drift_off}. In contrast, using a neural network approach to solve the index-1 problem can yield more accurate results for the algebraic constraints of high-index problems. Fig. \ref{fig:particle_drift_off} presents the drift-off error curves for both the DAE-KAN and traditional PINNs models when solving equation \eqref{4.2}. The error curves in the first column show the drift-off errors obtained by substituting the solutions for the differential variables \( u_1\), \(u_2\), \(z_1\), \(z_2\), and \(\lambda\) from both models into the corresponding algebraic equations for index-1, index-2, and index-3 formulations, i.e., \eqref{4.4}, \eqref{4.3}, and the final equation of \eqref{4.2}. Similarly, the error curves in the second and third columns display the corresponding drift-off errors. It can be observed that, regardless of whether the traditional PINNs model or the DAE-KAN model is used, the drift-off errors are significantly smaller when the solutions for the differential variables are substituted into the algebraic equations for index-1, index-2, and index-3 formulations, compared to those produced by traditional numerical methods.

\subsection{Dynamics of a Two-Link Planar Robot Arm on a Predefined Trajectory}

\begin{figure}[h]
    \centering
    \includegraphics[width=0.7\linewidth]{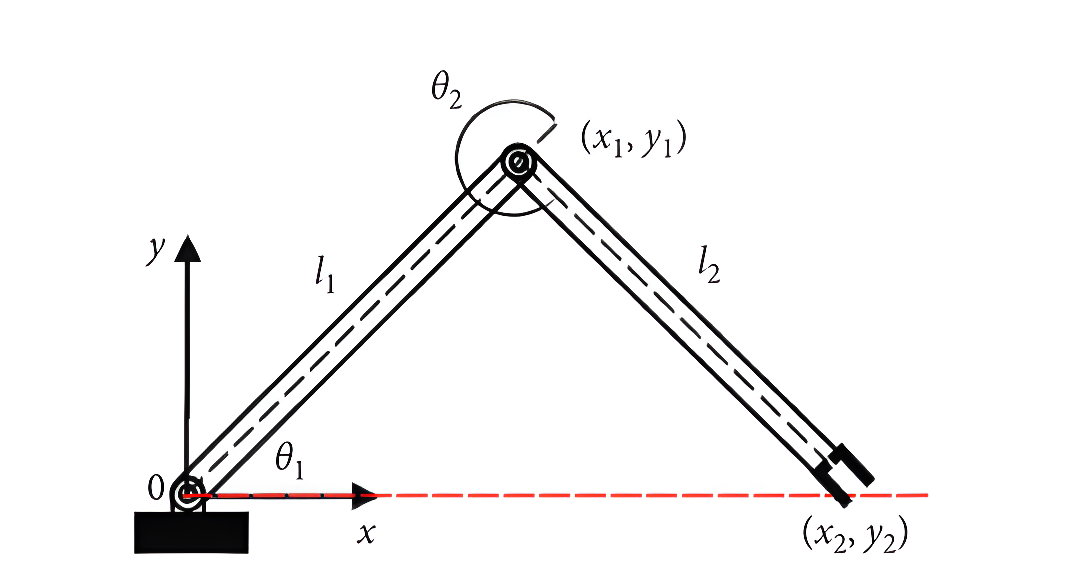}
    \caption{Schematic diagram of a robot arm.}
    \label{fig:robot_arm}
\end{figure}

As shown in Fig. \ref{fig:robot_arm}, the robot arm \cite{Ascher1997} consists of two rigid links. The arm consists of two rigid links. The first link (the left link) has its left end fixed at the origin of the $xy$-plane, but it can rotate. The angle of rotation of this end is measured from the $x$-axis and is denoted by $\theta_1$. The right end of the first link is connected to the left end of the second link (the right link) at $(x_1, y_1)$. The second link can rotate within the $xy$-plane, and its rotation angle is denoted by $\theta_2$. The masses and lengths of the two links are denoted by $m_i$ and $l_i$, where $i = 1, 2$.

The coordinates of the intermediate joint of the robot arm $(x_1, y_1)$ are given by:
\begin{equation}
\left\{
\begin{aligned}
    x_1 &= l_1 \cos\theta_1, \\
    y_1 &= l_1 \sin\theta_1,
\end{aligned}
\right.
\tag{4.7}
\end{equation}
The coordinates of the end effector of the robot arm $(x_2, y_2)$ are given by:
\begin{equation}
\left\{
\begin{aligned}
    x_2 &= x_1 + l_2 \cos(\theta_1 + \theta_2), \\
    y_2 &= y_1 + l_2 \sin(\theta_1 + \theta_2),
\end{aligned}
\right.
\tag{4.8}
\end{equation}
If the end effector of the robot arm is constrained to move along the $x$-axis, then $y_2 = 0$, and the following constraint on the angles $\theta_1$ and $\theta_2$ is obtained:
\begin{equation}
    l_1 \sin\theta_1 + l_2 \sin(\theta_1 + \theta_2) = 0.
\tag{4.9}
\end{equation}
Thus, Fig. \ref{fig:robot_arm} can be described using equation \eqref{2.2}. For this example of the robot arm, the mass matrix is:
\begin{align}   
M(\theta_1,\theta_2)= \left (\begin{matrix}
m_{1}\frac{l_{1}^{2}}{3}+m_{2}\left(l_{1}^{2}+\frac{l_{2}^{2}}{3}+l_{1}l_{2}\cos\theta_{2}\right) & m_{2}\left(\frac{l_{2}^{2}}{3}+0.5l_{1}l_{2}\cos\theta_{2}\right) \\
m_{2}\left(\frac{l_{2}^{2}}{3}+0.5l_{1}l_{2}\cos\theta_{2}\right) & m_{2}\frac{l_{2}^{2}}{3}
\end{matrix} \right ).\tag{4.10}
\end{align}
The applied force term is:
\begin{align}
f(\theta _{1}, \theta _{2}, \frac{{d}\theta _{1}}{{d}t}, \frac{{d}\theta _{2}}{{d}t}, t)=\left (\begin{matrix} (l_{1}\cos \theta _{1}+l_{2}\cos (\theta _{1}+\theta _{2}))({d}\theta _{1}/{{d}t})-3\theta _{1}\\ l_{2}\cos (\theta _{1}+\theta _{2})({d}\theta _{1}/{{d}t})+(1-1.5\cos \theta _{2})\theta_1
\end{matrix} \right ).
\tag{4.11}
\end{align}
The constraint function is
\begin{align} 
g(\theta _{1}, \theta _{2}, t)=l_{1}\sin \theta _{1}+l_{2}\sin (\theta _{1}+\theta _{2}). \tag{4.12}
\end{align}

In this numerical simulation, we set $l_{1}=l_{2}=1$ and $m_{1}=m_{2}=3$. Using the DAE systems \eqref{2.2} with the notation where $n_{u}=2, n_{\lambda}=1, u=(u_{1}, u_{2})^{T}=(\theta _{1}, \theta _{2})^{T}$, and $v={du}/{dt}$, we have
\begin{align} 
M(u)=\left (\begin{matrix} 5+3\cos u_{2}&1+1.5\cos u_{2}\\ 1+1.5\cos u_{2}&1\end{matrix}, \right ) 
\tag{4.13}
\end{align}
and
\begin{align} 
f(u, v, t)=\left (\begin{matrix} (\cos u_{1}+\cos (u_{1}+u_{2}))v_{1}-3u_{1}\\ \cos (u_{1}+u_{2})v_{1}+(1-1.5\cos u_{2})u_{1}\end{matrix} \right ). \tag{4.14}
\end{align}
the constraint function becomes
\begin{align} 
\label{3.16}
g(u, t)=\sin u_{1}+\sin (u_{1}+u_{2}), \tag{4.15}
\end{align}
differentiating the above constraint function once and simplifying, we obtain the constraint function corresponding to the index-2 form of the system:
\begin{align} 
\label{3.17}
g(u,v,t) = \cos(u_1)v_1 + \cos(u_1+u_2)(v_1+v_2) ,\tag{4.16}
\end{align}
we differentiate the above constraint function \eqref{3.17} again and simplify, we obtain the constraint function corresponding to the index-1 form of the system:
\begin{align} 
\label{3.18}
g(u,v,t) = -\sin(u_1)v_1^2+\cos(u_1)v_1’ + (v_1’ + v_2’)\cos(u_1+u_2)-\sin(u_1+u_2)(v_1+v_2)^2.\tag{4.17}
\end{align}
\begin{figure}[h]
    \centering
    \includegraphics[width=1\linewidth]{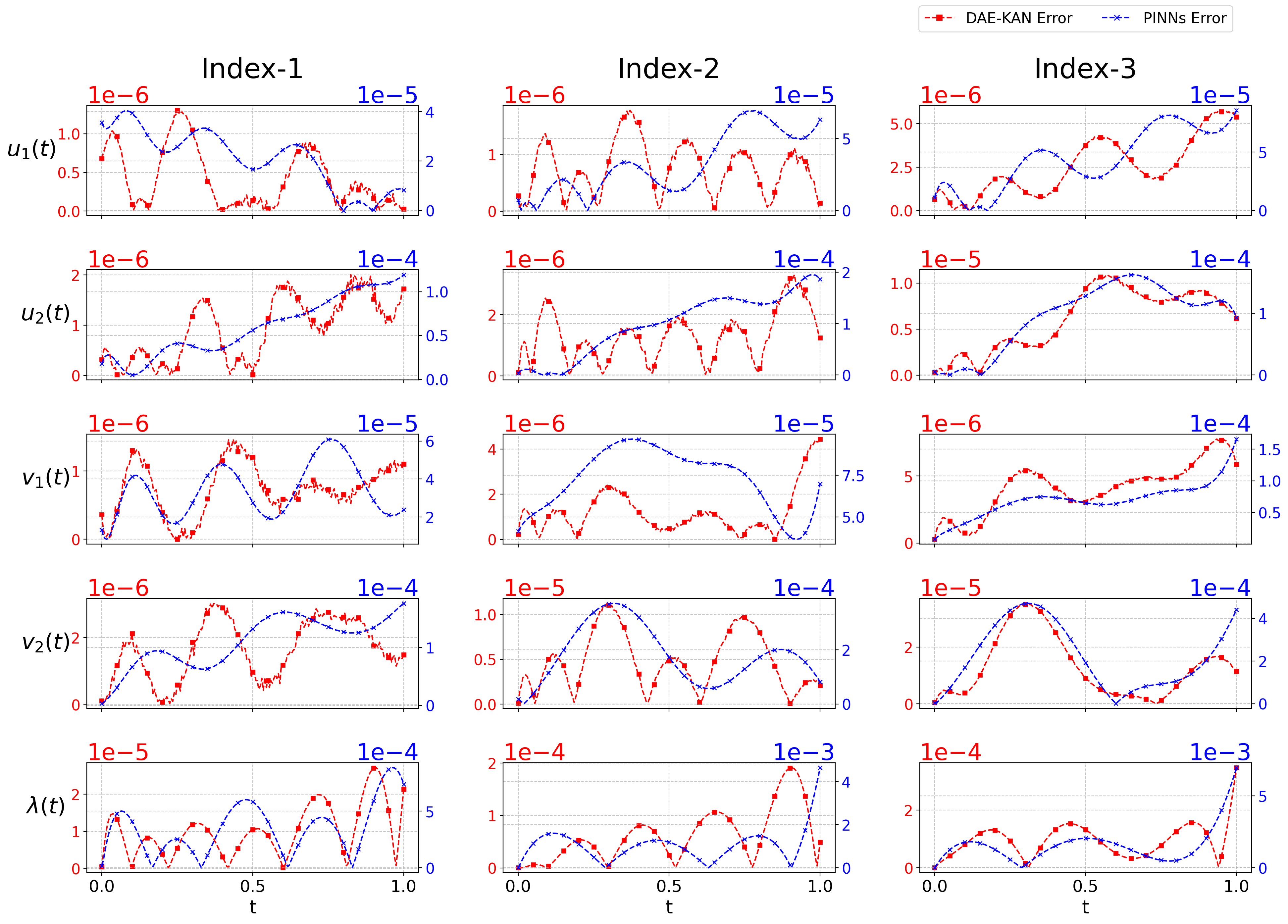}
    \caption{The \(\text{AE}\) curves of the predicted solution and the exact solution obtained by using the traditional PINNs and DAE-KAN to solve the corresponding index-1, index-2, and index-3 forms of \eqref{2.2}, respectively. The green line represents DAE-KAN, and the blue line represents PINNs.}
    \label{fig:robot_Absolute_errors}
\end{figure}
The exact solution of the DAEs corresponding to this robot arm problem is given by $ u(t) = (\sin t, -2\sin t)^T ,  v(t) = (\cos t, -2\cos t)^T $ and $\lambda(t) = \cos t $. This solution corresponds to the initial conditions:
\begin{align} 
u(0) = (0, 0)^T, v(0) = (1, -2)^T,\lambda(0) = 1. \tag{4.18}
\end{align}

Firstly, the differential KAN network of the DAE-KAN model is constructed as a three-layer KAN network with dimensions $[1,4,4,4]$; the algebraic KAN network is constructed as a three-layer KAN network with dimensions $[1,2,2,1]$. Similarly, based on the Loss function \eqref{2.16}, we set the number of initial points to 1, i.e., $N_i=1$, and the number of residual points to 200, i.e., $N_F=200$. These 200 points are uniformly sampled within the interval $t\in[0,1]$, forming our training dataset consisting of initial and residual points, totaling 201 points. The training process utilizes the LBFGS algorithm over 20,000 epochs. The absolute errors between predicted and exact solutions are depicted by the green curve in Fig. \ref{fig:robot_Absolute_errors}.

This paper also compares the traditional PINNs model \cite{PINN} and the DAE-KAN model in training and prediction for forms corresponding to equation \eqref{2.2}'s index-1, index-2, and index-3. The traditional PINNs model employs a deep neural network with 5 hidden layers, each containing 80 neurons using the $Tanh$ function as an activation function. The \(\text{AE}\) curves for both network models are shown in Fig. \ref{fig:robot_Absolute_errors}: the blue curve represents the \(\text{AE}\) of the traditional PINNs model, while the red curve represents that of the DAE-KAN model. From Fig. \ref{fig:robot_Absolute_errors}, it can be observed that the traditional PINNs model achieves an \(\text{AE}\) precision of $10^{-4} \sim 10^{-5}$ for solving the differential variables $u_1$, $u_2$, $v_1$, $v_2$ in the form of equation \eqref{2.2}, and $10^{-3}$ for the algebraic variable $\lambda$. In contrast, the DAE-KAN model achieves a higher precision with \(\text{AE}\) of $10^{-5} \sim 10^{-6}$ for $u_1$, $v_1$, $v_2$, surpassing the traditional PINNs model by an order of magnitude for $u_1$ and achieving two orders of magnitude improvement for $u_2$; similarly, the \(\text{AE}\) precision for the algebraic variable $\lambda$ is improved by one order of magnitude, reaching $10^{-4}$.

Replacing the algebraic equation of \eqref{2.2} with \eqref{3.17} results in its index-2 form. The traditional PINNs model predicts \(\text{AE}\) precisions of $10^{-4} \sim 10^{-5}$ for the differential variables $u_1$, $u_2$, $v_1$, $v_2$ and $10^{-3}$ for the algebraic variable $v$. Meanwhile, the DAE-KAN model predicts \(\text{AE}\) precisions of $10^{-5} \sim 10^{-6}$ for $u_1$, $u_2$, $v_1$, $v_2$, achieving two orders of magnitude improvement for $v_1$ compared to the traditional PINNs model, one order of magnitude improvement for $u_1$, $u_2$, $v_2$, and $\lambda$'s \(\text{AE}\) precision is also improved by one order of magnitude, reaching $10^{-4}$.Replacing the algebraic equation of \eqref{2.2} with \eqref{3.18} results in its index-1 form, where it can be seen that both in terms of differential and algebraic variables, DAE-KAN outperforms traditional PINNs, exhibiting lower \(\text{AE}\).

According to Table \ref{tb2}, both models compare the predicted solutions and exact solutions of equation \eqref{2.2} corresponding to index-2 and index-3 forms in terms of \(\text{RE}\). In the context of index-3 form of  \eqref{2.2}, the DAE-KAN model's overall maximum \(\text{RE}\) for predicting differential variables is $9.85 \times 10^{-6}$, surpassing the traditional PINNs model by two orders of magnitude. Moreover, in the prediction of algebraic variables, the DAE-KAN model also outperforms the traditional PINNs model. In the index-2 form of  \eqref{2.2}, the DAE-KAN model's overall maximum \(\text{RE}\) is superior to that of the traditional PINNs model, with an improvement of up to two orders of magnitude in the prediction accuracy of differential variables $u_2$ and $v_2$. The comparison of \(\text{RE}\) in the index-1 form also shows that DAE-KAN similarly outperforms traditional PINNs.

\begin{figure}[h]
    \centering
    \includegraphics[width=1\linewidth]{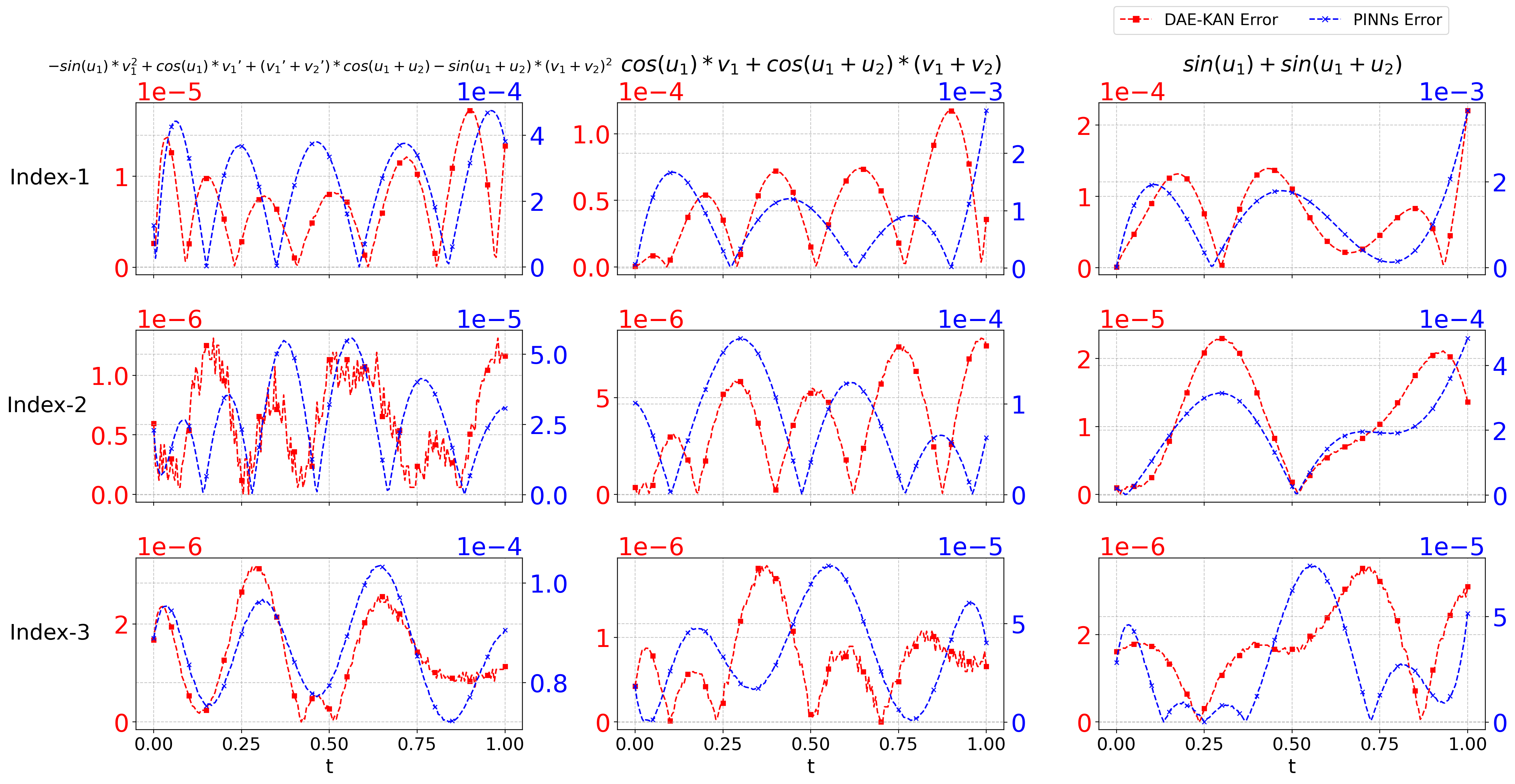}
    \caption{The predicted solutions in the form of \eqref{2.2} index-1 index-2 and index-3 obtained by PINNs and DAE-KAN are respectively put into the constraint equations in the form of \eqref{2.2} index-1 index-2 and index-3. That is, the Eqs \eqref{3.18} \eqref{3.17} and \eqref{3.16}, and finally the drift-off error curve is obtained. Where the green line represents the drift-off error of DAE-KAN and the blue line represents the drift-off error of PINNs.}
    \label{fig:robot_drift_off}
\end{figure}
\begin{table}[width=.9\linewidth,cols=6,pos=h]
\renewcommand{\arraystretch}{1.5}
\caption{Comparison of \(\text{RE}\) for different models and variables.}\label{tb2}
\begin{tabular*}{\tblwidth}{@{} LLLLLL@{} }
\toprule
Model & $u_1(t)$ & $u_2(t)$ & $v_1(t)$ & $v_2(t)$ & $\lambda(t)$ \\
\midrule
DAE-KAN(index-3) & \textbf{5.75e-06} & \textbf{6.64e-06} & \textbf{5.34e-06} & \textbf{9.85e-06} & \textbf{1.26e-04} \\
PINNs(index-3) & 9.70e-05 & 1.04e-04 & 8.73e-05 & 1.57e-04 & 2.24e-03 \\
DAE-KAN(index-2) & \textbf{1.67e-06} & \textbf{1.46e-06} & \textbf{1.79e-06} & \textbf{3.33e-06} & \textbf{9.27e-05} \\
PINNs(index-2) & 7.44e-05 & 1.09e-04 & 8.76e-05 & 1.23e-04 & 1.47e-03 \\
DAE-KAN(index-1) & \textbf{1.07e-06} & \textbf{1.06e-06} & \textbf{9.64e-07} & \textbf{1.09e-06} & \textbf{1.41e-05} \\
PINNs(index-1) & 4.60e-05 & 6.53e-05 & 4.27e-05 & 6.85e-05 & 4.86e-04 \\
\bottomrule
\end{tabular*}
\end{table}
Fig. \ref{fig:robot_drift_off} presents the drift-off error curves obtained by solving equation \eqref{2.2} using both the DAE-KAN model and the traditional PINNs model. In the first column, the error curves represent the solutions $u_1$, $u_2$, $v_1$, $v_2$, $\lambda$ obtained by solving the index-1 form of the equation using both models, which are then substituted into the algebraic equations corresponding to the index-2 and index-3 forms of the DAE systems, i.e., \eqref{3.17} and \eqref{3.16}. Similarly, the second and third columns follow the same procedure.
It can be observed that for both the traditional PINNs model and the DAE-KAN model, the drift-off errors remain relatively small when the solutions obtained from solving the index-1, index-2, and index-3 forms are \text{RE}-substituted into the algebraic equations corresponding to the respective forms. This demonstrates the potential of neural network methods in handling drift-off errors effectively.

The above numerical results indicate that, after sufficient iterations, DAE-KAN can gradually learn the physical laws of the real system, more effectively capture the relationships between the different variables of the DAE systems, and achieve higher prediction accuracy and stability compared to the traditional PINNs \cite{PINN} solution model. This addresses the issue of poor accuracy in existing neural network solutions for DAE equations, explores the ability of neural networks to overcome drift-off errors, and provides convenience for future engineering research.

\section{Conclusion}\label{sec5}

This work presents DAE-KAN, a deep learning framework that integrates Kolmogorov-Arnold Networks (KANs) with the physics-informed neural network approach. This innovative combination of data-driven techniques enhances the ability to model high-index differential-algebraic systems in scientific machine learning. By utilizing trainable B-spline activation functions and a dual-network structure, where separate KAN networks predict differential and algebraic variables, DAE-KAN significantly improves solution accuracy for high-index DAEs. Numerical experiments demonstrate that DAE-KAN reduces the absolute errors of both differential and algebraic variables by 1-2 orders of magnitude compared to traditional PINNs when solving DAEs with index-1 to index-3, and it excels in mitigating drift-off errors in the constraint conditions. DAE-KAN offers a promising new approach for efficiently modeling complex engineering systems. Future research could explore the adaptability of the KAN network to a broader range of DAE types, investigate more complex physical systems, and integrate DAE-KAN with existing numerical methods to further enhance solution accuracy.

\bibliographystyle{elsarticle-num-names}
\bibliography{references}

\begin{thebibliography}{30}
\expandafter\ifx\csname natexlab\endcsname\relax\def\natexlab#1{#1}\fi
\providecommand{\url}[1]{\texttt{#1}}
\providecommand{\href}[2]{#2}
\providecommand{\path}[1]{#1}
\providecommand{\DOIprefix}{doi:}
\providecommand{\ArXivprefix}{arXiv:}
\providecommand{\URLprefix}{URL: }
\providecommand{\Pubmedprefix}{pmid:}
\providecommand{\doi}[1]{\href{http://dx.doi.org/#1}{\path{#1}}}
\providecommand{\Pubmed}[1]{\href{pmid:#1}{\path{#1}}}
\providecommand{\bibinfo}[2]{#2}
\ifx\xfnm\relax \def\xfnm[#1]{\unskip,\space#1}\fi
\bibitem[{Brenan et~al.(1995)Brenan, Campbell, and Petzold}]{Brenan1996}
\bibinfo{author}{K.~E. Brenan}, \bibinfo{author}{S.~L. Campbell}, \bibinfo{author}{L.~R. Petzold}, \bibinfo{title}{Numerical solution of initial-value problems in differential-algebraic equations}, \bibinfo{publisher}{SIAM}, \bibinfo{year}{1995}.
\bibitem[{Gear(1971)}]{Gear1971}
\bibinfo{author}{C.~Gear},
\newblock \bibinfo{title}{Simultaneous numerical solution of differential-algebraic equations},
\newblock \bibinfo{journal}{IEEE transactions on circuit theory} \bibinfo{volume}{18} (\bibinfo{year}{1971}) \bibinfo{pages}{89--95}.
\bibitem[{Biegler(2007)}]{Biegler2007}
\bibinfo{author}{L.~T. Biegler},
\newblock \bibinfo{title}{An overview of simultaneous strategies for dynamic optimization},
\newblock \bibinfo{journal}{Chemical Engineering and Processing: Process Intensification} \bibinfo{volume}{46} (\bibinfo{year}{2007}) \bibinfo{pages}{1043--1053}.
\bibitem[{Geilinger et~al.(2020)Geilinger, Hahn, Zehnder, B{\"a}cher, Thomaszewski, and Coros}]{robotic}
\bibinfo{author}{M.~Geilinger}, \bibinfo{author}{D.~Hahn}, \bibinfo{author}{J.~Zehnder}, \bibinfo{author}{M.~B{\"a}cher}, \bibinfo{author}{B.~Thomaszewski}, \bibinfo{author}{S.~Coros},
\newblock \bibinfo{title}{Add: Analytically differentiable dynamics for multi-body systems with frictional contact},
\newblock \bibinfo{journal}{ACM Transactions on Graphics (TOG)} \bibinfo{volume}{39} (\bibinfo{year}{2020}) \bibinfo{pages}{1--15}.
\bibitem[{Simeon et~al.(1991)Simeon, F{\"u}hrer, and Rentrop}]{Simeon1991}
\bibinfo{author}{B.~Simeon}, \bibinfo{author}{C.~F{\"u}hrer}, \bibinfo{author}{P.~Rentrop},
\newblock \bibinfo{title}{Differential-algebraic equations in vehicle system dynamics},
\newblock \bibinfo{journal}{Surv. Math. Ind.}  (\bibinfo{year}{1991}) \bibinfo{pages}{1--37}.
\bibitem[{Ascher and Petzold(1991)}]{AscherPetzold1991}
\bibinfo{author}{U.~M. Ascher}, \bibinfo{author}{L.~R. Petzold},
\newblock \bibinfo{title}{Projected implicit {R}unge--{K}utta methods for differential-algebraic equations},
\newblock \bibinfo{journal}{SIAM Journal on Numerical Analysis} \bibinfo{volume}{28} (\bibinfo{year}{1991}) \bibinfo{pages}{1097--1120}.
\bibitem[{Cash(2000)}]{Cash2000}
\bibinfo{author}{J.~Cash},
\newblock \bibinfo{title}{Modified extended backward differentiation formulae for the numerical solution of stiff initial value problems in {ODE}s and {DAE}s},
\newblock \bibinfo{journal}{Journal of Computational and Applied Mathematics} \bibinfo{volume}{125} (\bibinfo{year}{2000}) \bibinfo{pages}{117--130}.
\bibitem[{Jay and Negrut(2009)}]{jay2009}
\bibinfo{author}{O.~L. Jay}, \bibinfo{author}{D.~Negrut},
\newblock \bibinfo{title}{A second order extension of the generalized--$\alpha$ method for constrained systems in mechanics},
\newblock in: \bibinfo{booktitle}{Multibody Dynamics: Computational Methods and Applications}, \bibinfo{publisher}{Springer}, \bibinfo{year}{2009}, pp. \bibinfo{pages}{143--158}.
\bibitem[{Liu et~al.(2017)Liu, Chen, and Liu}]{Liu2017}
\bibinfo{author}{C.-S. Liu}, \bibinfo{author}{W.~Chen}, \bibinfo{author}{L.-W. Liu},
\newblock \bibinfo{title}{Solving mechanical systems with nonholonomic constraints by a lie-group differential algebraic equations method},
\newblock \bibinfo{journal}{Journal of Engineering Mechanics} \bibinfo{volume}{143} (\bibinfo{year}{2017}) \bibinfo{pages}{04017097}.
\bibitem[{Tang and Lu(2023)}]{Tang2023}
\bibinfo{author}{J.~Tang}, \bibinfo{author}{J.~Lu},
\newblock \bibinfo{title}{Modified extended lie-group method for hessenberg differential algebraic equations with index-3},
\newblock \bibinfo{journal}{Mathematics} \bibinfo{volume}{11} (\bibinfo{year}{2023}) \bibinfo{pages}{2360}.
\bibitem[{Gear and Petzold(1984)}]{Gear1984}
\bibinfo{author}{C.~W. Gear}, \bibinfo{author}{L.~R. Petzold},
\newblock \bibinfo{title}{{ODE} methods for the solution of differential/algebraic systems},
\newblock \bibinfo{journal}{SIAM Journal on Numerical analysis} \bibinfo{volume}{21} (\bibinfo{year}{1984}) \bibinfo{pages}{716--728}.
\bibitem[{Yang et~al.(2022)Yang, Wu, and Reid}]{yang2022index}
\bibinfo{author}{W.~Yang}, \bibinfo{author}{W.~Wu}, \bibinfo{author}{G.~Reid},
\newblock \bibinfo{title}{Index reduction for degenerated differential-algebraic equations by embedding},
\newblock \bibinfo{journal}{arXiv preprint arXiv:2210.16707}  (\bibinfo{year}{2022}).
\bibitem[{Qin et~al.(2016)Qin, Tang, Feng, Bachmann, and Fritzson}]{Tang2016}
\bibinfo{author}{X.~Qin}, \bibinfo{author}{J.~Tang}, \bibinfo{author}{Y.~Feng}, \bibinfo{author}{B.~Bachmann}, \bibinfo{author}{P.~Fritzson},
\newblock \bibinfo{title}{Efficient index reduction algorithm for large scale systems of differential algebraic equations},
\newblock \bibinfo{journal}{Applied Mathematics and Computation} \bibinfo{volume}{277} (\bibinfo{year}{2016}) \bibinfo{pages}{10--22}.
\bibitem[{Tang and Rao(2020)}]{Tang2020}
\bibinfo{author}{J.~Tang}, \bibinfo{author}{Y.~Rao},
\newblock \bibinfo{title}{A new block structural index reduction approach for large-scale differential algebraic equations},
\newblock \bibinfo{journal}{Mathematics} \bibinfo{volume}{8} (\bibinfo{year}{2020}) \bibinfo{pages}{2057}.
\bibitem[{Raissi et~al.(2019)Raissi, Perdikaris, and Karniadakis}]{PINN}
\bibinfo{author}{M.~Raissi}, \bibinfo{author}{P.~Perdikaris}, \bibinfo{author}{G.~E. Karniadakis},
\newblock \bibinfo{title}{Physics-informed neural networks: A deep learning framework for solving forward and inverse problems involving nonlinear partial differential equations},
\newblock \bibinfo{journal}{Journal of Computational physics} \bibinfo{volume}{378} (\bibinfo{year}{2019}) \bibinfo{pages}{686--707}.
\bibitem[{Yu et~al.(2022)Yu, Lu, Meng, and Karniadakis}]{Yu2022}
\bibinfo{author}{J.~Yu}, \bibinfo{author}{L.~Lu}, \bibinfo{author}{X.~Meng}, \bibinfo{author}{G.~E. Karniadakis},
\newblock \bibinfo{title}{Gradient-enhanced physics-informed neural networks for forward and inverse {PDE} problems},
\newblock \bibinfo{journal}{Computer Methods in Applied Mechanics and Engineering} \bibinfo{volume}{393} (\bibinfo{year}{2022}) \bibinfo{pages}{114823}.
\bibitem[{Wang et~al.(2021)Wang, Teng, and Perdikaris}]{Wang2021}
\bibinfo{author}{S.~Wang}, \bibinfo{author}{Y.~Teng}, \bibinfo{author}{P.~Perdikaris},
\newblock \bibinfo{title}{Understanding and mitigating gradient flow pathologies in physics-informed neural networks},
\newblock \bibinfo{journal}{SIAM Journal on Scientific Computing} \bibinfo{volume}{43} (\bibinfo{year}{2021}) \bibinfo{pages}{A3055--A3081}.
\bibitem[{Kozlov and Tiumentsev(2018)}]{Kozlov2018}
\bibinfo{author}{D.~S. Kozlov}, \bibinfo{author}{Y.~V. Tiumentsev},
\newblock \bibinfo{title}{Neural network based semi-empirical models for dynamical systems represented by differential-algebraic equations of index 2},
\newblock \bibinfo{journal}{Procedia computer science} \bibinfo{volume}{123} (\bibinfo{year}{2018}) \bibinfo{pages}{252--257}.
\bibitem[{Moya and Lin(2023)}]{DAE-PINN}
\bibinfo{author}{C.~Moya}, \bibinfo{author}{G.~Lin},
\newblock \bibinfo{title}{{DAE-PINN}: a physics-informed neural network model for simulating differential algebraic equations with application to power networks},
\newblock \bibinfo{journal}{Neural Computing and Applications} \bibinfo{volume}{35} (\bibinfo{year}{2023}) \bibinfo{pages}{3789--3804}.
\bibitem[{Chen et~al.(2024)Chen, Tang, Yan, Lai, Liang, Lu, and Yang}]{Radau-PINN}
\bibinfo{author}{J.~Chen}, \bibinfo{author}{J.~Tang}, \bibinfo{author}{M.~Yan}, \bibinfo{author}{S.~Lai}, \bibinfo{author}{K.~Liang}, \bibinfo{author}{J.~Lu}, \bibinfo{author}{W.~Yang},
\newblock \bibinfo{title}{Physics-informed neural networks for solving high-index differential-algebraic equation systems based on radau methods},
\newblock \bibinfo{journal}{International Journal of Intelligent Systems} \bibinfo{volume}{2024} (\bibinfo{year}{2024}) \bibinfo{pages}{6641674}.
\bibitem[{Hairer and Wanner(1999)}]{Hairer2015}
\bibinfo{author}{E.~Hairer}, \bibinfo{author}{G.~Wanner},
\newblock \bibinfo{title}{Stiff differential equations solved by {R}adau methods},
\newblock \bibinfo{journal}{Journal of Computational and Applied Mathematics} \bibinfo{volume}{111} (\bibinfo{year}{1999}) \bibinfo{pages}{93--111}.
\bibitem[{Liu et~al.(2024)Liu, Wang, Vaidya, Ruehle, Halverson, Solja{\v{c}}i{\'c}, Hou, and Tegmark}]{KAN}
\bibinfo{author}{Z.~Liu}, \bibinfo{author}{Y.~Wang}, \bibinfo{author}{S.~Vaidya}, \bibinfo{author}{F.~Ruehle}, \bibinfo{author}{J.~Halverson}, \bibinfo{author}{M.~Solja{\v{c}}i{\'c}}, \bibinfo{author}{T.~Y. Hou}, \bibinfo{author}{M.~Tegmark},
\newblock \bibinfo{title}{{KAN}: Kolmogorov-arnold networks},
\newblock \bibinfo{journal}{arXiv preprint arXiv:2404.19756}  (\bibinfo{year}{2024}).
\bibitem[{Kolmogorov(1961)}]{Kolmogorov1961}
\bibinfo{author}{A.~N. Kolmogorov}, \bibinfo{title}{On the representation of continuous functions of several variables by superpositions of continuous functions of a smaller number of variables}, \bibinfo{publisher}{American Mathematical Society}, \bibinfo{year}{1961}.
\bibitem[{Shabana(2020)}]{Shabana_2013}
\bibinfo{author}{A.~A. Shabana}, \bibinfo{title}{Dynamics of multibody systems}, \bibinfo{publisher}{Cambridge university press}, \bibinfo{year}{2020}.
\bibitem[{M{\"a}rz(2002)}]{Marz2002}
\bibinfo{author}{R.~M{\"a}rz},
\newblock \bibinfo{title}{The index of linear differential algebraic equations with properly stated leading terms},
\newblock \bibinfo{journal}{Results in Mathematics} \bibinfo{volume}{42} (\bibinfo{year}{2002}) \bibinfo{pages}{308--338}.
\bibitem[{Wanner and Hairer(1996)}]{hairer1991solving}
\bibinfo{author}{G.~Wanner}, \bibinfo{author}{E.~Hairer}, \bibinfo{title}{Solving ordinary differential equations II}, volume \bibinfo{volume}{375}, \bibinfo{publisher}{Springer Berlin Heidelberg New York}, \bibinfo{year}{1996}.
\bibitem[{Hornik et~al.(1989)Hornik, Stinchcombe, and White}]{Hornik1989}
\bibinfo{author}{K.~Hornik}, \bibinfo{author}{M.~Stinchcombe}, \bibinfo{author}{H.~White},
\newblock \bibinfo{title}{Multilayer feedforward networks are universal approximators},
\newblock \bibinfo{journal}{Neural networks} \bibinfo{volume}{2} (\bibinfo{year}{1989}) \bibinfo{pages}{359--366}.
\bibitem[{Li(2024)}]{li2024kolmogorov}
\bibinfo{author}{Z.~Li},
\newblock \bibinfo{title}{Kolmogorov-arnold networks are radial basis function networks},
\newblock \bibinfo{journal}{arXiv preprint arXiv:2405.06721}  (\bibinfo{year}{2024}).
\bibitem[{Benhammouda(2015)}]{Benhammouda2015}
\bibinfo{author}{B.~Benhammouda},
\newblock \bibinfo{title}{Solution of nonlinear higher-index hessenberg {DAE}s by adomian polynomials and differential transform method},
\newblock \bibinfo{journal}{SpringerPlus} \bibinfo{volume}{4} (\bibinfo{year}{2015}) \bibinfo{pages}{648}.
\bibitem[{Ascher and Lin(1997)}]{Ascher1997}
\bibinfo{author}{U.~Ascher}, \bibinfo{author}{P.~Lin},
\newblock \bibinfo{title}{Sequential regularization methods for nonlinear higher-index {DAE}s},
\newblock \bibinfo{journal}{SIAM Journal on Scientific Computing} \bibinfo{volume}{18} (\bibinfo{year}{1997}) \bibinfo{pages}{160--181}.

\end{thebibliography}

\end{document}